\documentclass{article} %
\usepackage[dvipsnames]{xcolor}
\usepackage{iclr2024_conference,times}
\iclrfinalcopy

\usepackage{amsmath,amsfonts,bm}

\def\eqref#1{equation~\ref{#1}}

\def\1{\bm{1}}

\DeclareMathAlphabet{\mathsfit}{\encodingdefault}{\sfdefault}{m}{sl}
\SetMathAlphabet{\mathsfit}{bold}{\encodingdefault}{\sfdefault}{bx}{n}

\usepackage{hyperref}
\usepackage{url}

\usepackage{booktabs} %
\usepackage{bbm}
\usepackage{multirow}
\usepackage{listings}
\usepackage{siunitx}
\usepackage{CJKutf8}
\usepackage{times}
\usepackage{latexsym}
\usepackage{microtype}
\usepackage{natbib}
\usepackage{booktabs}
\usepackage{appendix}
\usepackage{tikz}
\usepackage{tikz-dependency}
\usepackage{amsmath}
\usepackage{amsfonts}
\usepackage{graphicx}
\usepackage{amsthm}
\usepackage{mathtools}
\usepackage{multirow}
\usepackage{url}
\usepackage{color}

\usepackage{subfig}
\usepackage{dsfont}
\usepackage[ruled,vlined,noend]{algorithm2e}
\usepackage{wrapfig}

\usepackage[noabbrev,capitalize]{cleveref} %

\crefname{equation}{equation}{equations}   %
\crefname{footnote}{footnote}{footnotes}   %
\crefname{line}{line}{lines}               %
\crefname{section}{\S}{\S\S}
\Crefname{section}{\S}{\S\S}    %
\crefformat{section}{#2\S#1#3}  %
\Crefformat{section}{#2\S#1#3}
\crefrangeformat{section}{\S\S#3#1#4--#5#2#6}
\Crefrangeformat{section}{\S\S#3#1#4--#5#2#6}

\newcommand{\exper}[1]{\textsc{#1}}

\newcommand{\plm}{p_\exper{LM}}
\newcommand{\xx}{x}
\newcommand{\GG}{g}
\newcommand{\VV}{v}
\newcommand{\yy}{y}
\newcommand{\Gxx}{\xx_\textsc{G}}
\newcommand{\Gyy}{\yy_\textsc{G}}
\newcommand{\Dxx}{\xx_\textsc{V}}
\newcommand{\Dyy}{\yy_\textsc{V}}
\newcommand{\tempg}{\text{Temp}_\textsc{G}}
\newcommand{\tempd}{\text{Temp}_\textsc{V}}

\newcommand{\consistency}{c}

\newcommand{\logp}{\log p_\theta}

\newcommand{\GyyColor}[1]{\color{ProcessBlue}{#1}}
\newcommand{\DyyColor}[1]{\color{LimeGreen}{#1}}

\newcommand{\xxColor}[1]{\color{YellowOrange}{#1}}

\newcommand{\ConsistencyFT}{\textsc{Consistency}} 
\newcommand{\CTRL}{\textsc{CC-FT}} 
\newcommand{\SelfTrain}{\textsc{SelfTrain}} 
\newcommand{\Alpaca}{\textsc{Alpaca-30B}} 
\newcommand{\AlpacaSmall}{\textsc{Alpaca-7B}}

\usepackage{times}

\usepackage[T1]{fontenc}

\usepackage[utf8]{inputenc}

\usepackage{microtype}

\title{Benchmarking and Improving \\
Generator-Validator Consistency of LMs}

\author{Xiang Lisa Li,~~ Vaishnavi Shrivastava, ~~ Siyan Li, ~~ Tatsunori Hashimoto, ~~ Percy Liang  \\Stanford University, ~~Columbia University \\ \texttt{\{xlisali, vaish1, thashim\}@stanford.edu, siyan.li@columbia.edu}\\ \texttt{pliang@cs.stanford.edu}}

\begin{document}
\maketitle

\begin{abstract}
As of September 2023, ChatGPT correctly answers ``what is 7+8'' with 15, but when asked ``7+8=15, True or False'' it responds with ``False''.
This inconsistency between \emph{generating} and \emph{validating} an answer is prevalent in language models (LMs) and erodes trust. 
In this paper, we propose a framework for measuring the consistency between generation and validation (which we call generator-validator consistency, or GV-consistency), finding that even GPT-4, a state-of-the-art LM, is GV-consistent only 76\% of the time.
To improve the consistency of LMs, we propose to finetune on the filtered generator and validator responses that are GV-consistent, and call this approach consistency fine-tuning. 
We find that this approach improves GV-consistency of Alpaca-30B from 60\% to 93\%, and the improvement extrapolates to unseen tasks and domains (e.g., GV-consistency for positive style transfers extrapolates to unseen styles like humor).
In addition to improving consistency, consistency fine-tuning improves both generator quality and validator accuracy without using any labeled data.
Evaluated across 6 tasks, including math questions, knowledge-intensive QA, and instruction following, our method improves the generator quality by 16\% and the validator accuracy by 6.3\% across all tasks.\footnote{\url{https://github.com/XiangLi1999/GV-consistency}}

\end{abstract}

\section{Introduction}

\begin{wrapfigure}{r}{0.4\linewidth}
    \centering
    \vspace{-0.5cm}
    \includegraphics[width=0.4\textwidth, page=1]{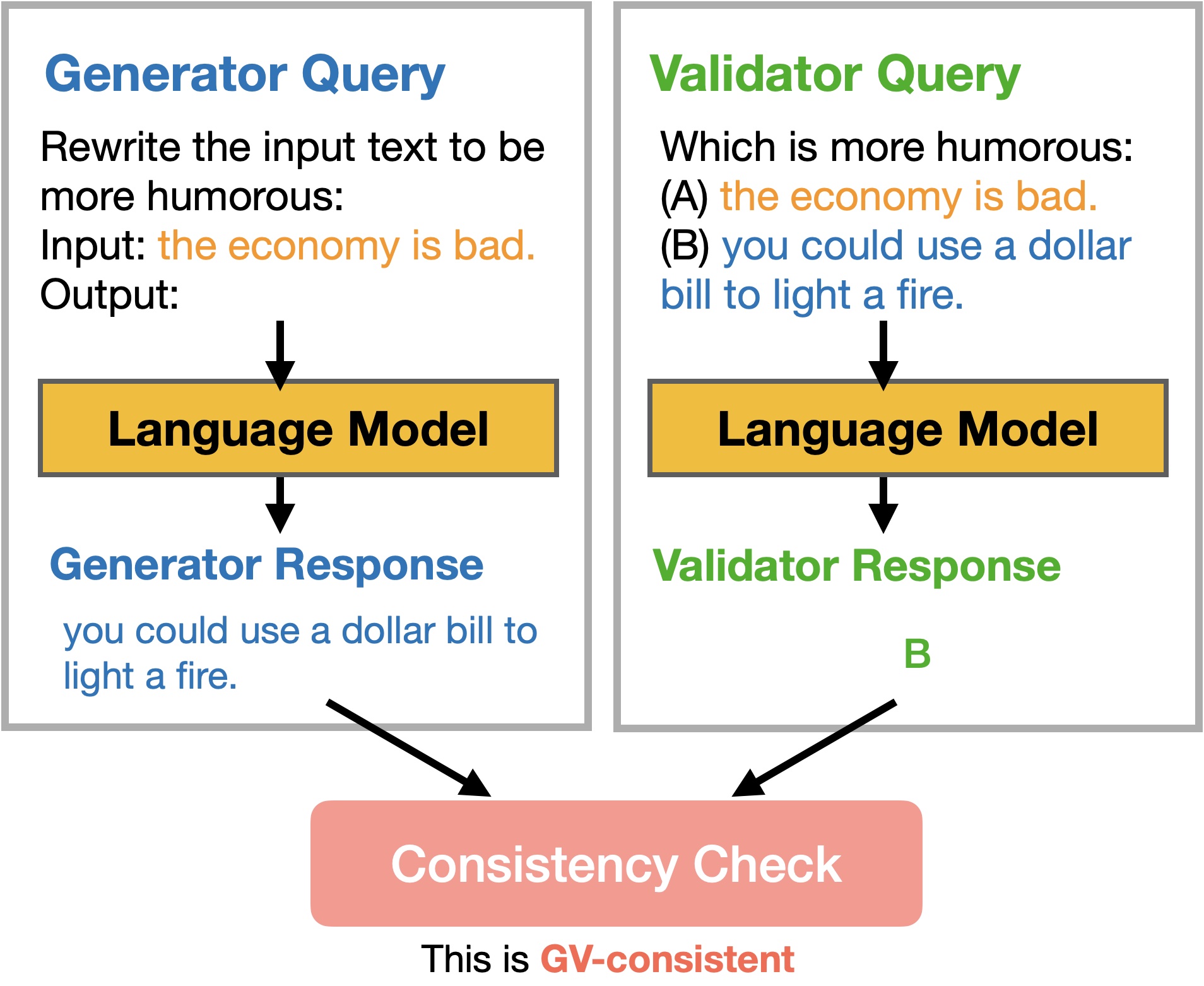} 
    \caption{\label{fig:fig1}
To measure generator-validator consistency, we prompt a LM with a generator query to produce a free-form answer. Then, we check if the same LM consistently responds to a corresponding validator query that asks if the generated answer is correct.
This example is GV-consistent because the validator confirms the generator response.
    }
\vspace{-1.0cm}
\end{wrapfigure}

Language models (LMs) can generate high-quality responses to task prompts; however, the same model can sometimes produce contradictory responses when validating its own answers.
For example, in September 2023, ChatGPT correctly responds to ``what is 7+8?'' with ``15'', but when prompted ``7+8=15, True or False'' it responds with ``False'' \footnote{\url{https://shorturl.at/ixPS5}}.
In this paper, we study a LM's consistency with respect to a \emph{generator} query that produces free-form text (e.g., ``what is 7+8?'') and its associated \emph{validator} query, which classifies whether the generator answer is correct or not (e.g., ``7+8=15, True or False?'').
A consistent LM that answers ``15'' to the generator query should also answer ``True'' to the validator query, and we call this consistency between generation and validation \emph{generator-validator consistency} or GV-consistency.

GV-consistency is a critical property for building trust in language models, and it can be applied to a broad range of tasks. 
Consistency of the generator and validator is key as both components form important use cases of language models: users often interact with LMs via generator queries, and prevalent approaches such as reinforcement learning from human feedback (RLHF) and classification tasks use validator queries as reward models and classifiers.
GV-consistency can also be applied to a broad range of tasks, as any open-ended generation task can also be formulated as a validator query that checks the correctness of the generator response.

In order to systematically assess GV-consistency of LMs, we propose a simple and scalable evaluation approach that relies on checking the consistency between carefully crafted generator and validator queries (\cref{sec:consis_eval}).
Our approach begins by prompting the LM with a generator query to solicit an answer to a question, and then prompting the same LM with a validator query to check whether the generated answer is correct. Simply asking the validator for a correctness judgment can fail, as the trivial baseline of always answering ``correct'' has perfect performance. Our work avoids this degeneracy by randomizing the labels corresponding to the consistent answer (\cref{ssec:GD-design}). 

\Cref{fig:fig1} shows an example validator query: which is more humorous? (A)  \texttt{\xxColor [original text]} or (B) \texttt{\GyyColor [generated text]}. A GV-consistent LM would respond to the validator query with the option corresponding to the generated text. Conversely, an inconsistent LM would choose the option corresponding to the original text, either due to the generator's failure to produce a more humorous text or the validator's inability to accurately gauge the humor level between the two sentences. 
We evaluated GV-consistency of GPT-4, GPT-3.5, text-davinci-003, and Alpaca-30B on math, QA, and instruction following tasks. We found that even state-of-the-art LMs struggle with GV-consistency: GPT-4 achieves only 76\% consistency and Alpaca-30B achieves only 60\%.

\begin{figure}
    \centering
    \includegraphics[width=0.99\textwidth, page=1]{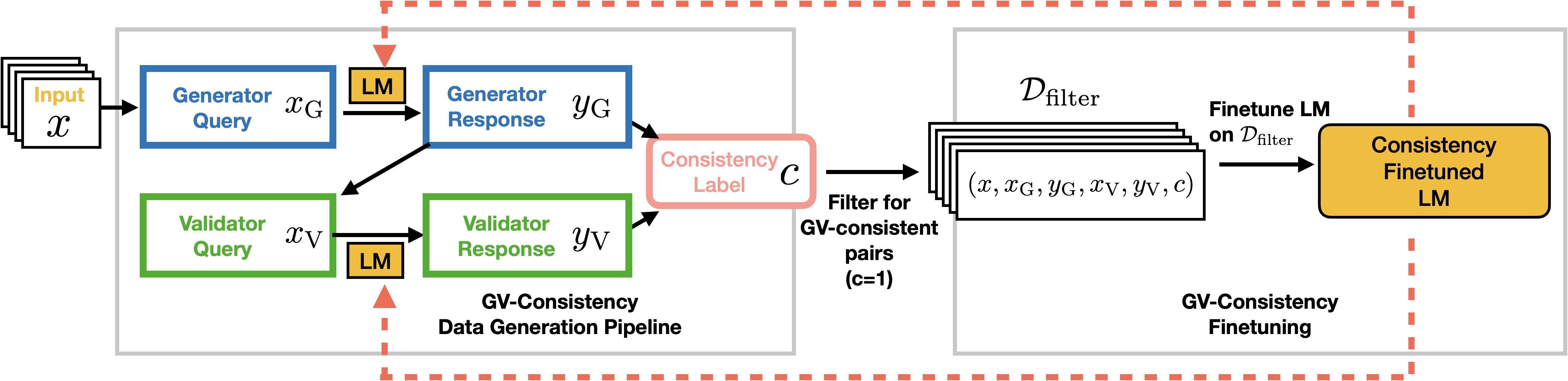}  
    \caption{\label{fig:fig2}
    GV-Consistency fine-tuning consists of two stages: the data generation stage, and the consistency fine-tuning stage. For the data generation stage, we collect the LM responses to generator queries and their associated validator queries.
    Next, we filter to only keep generator-validator response pairs that are consistent.
    Finally, we finetune the LM on the consistent pairs.
    This process can be iterated to further improve consistency.  
   \looseness=-1}
\vspace{-0.3cm}
\end{figure}

To improve GV-consistency, we propose a simple procedure called consistency fine-tuning, which consists of a data generation stage and a fine-tuning stage. As shown in \cref{fig:fig2}, 
given a generator and a validator prompt, we first query the generator to obtain the generator response, then query the validator to check the correctness of the generated response. 
We then filter the paired generator and discriminator responses to keep only the pairs that are GV-consistent. Finally, we finetune the LM to maximize the likelihood of the consistent pairs. Crucially, our approach only requires \emph{unlabeled data}. 
Moreover, this algorithm can be applied for multiple rounds: (1) generate the generator-validator data pairs using the newly fine-tuned LM, (2) finetune the LM on the consistent subset, and (3) repeat (as shown by the red arrows).

To evaluate consistency fine-tuning, We experiment on 6 tasks, ranging from classic NLP tasks (style transfer and QA) to arithmetic reasoning (arithmetic and plan arithmetic) and instruction-following (harmful question and prompt prioritization). Across all 6 tasks, we find that our consistency fine-tuning significantly improves the GV-consistency of Alpaca-30B from 60\% to 94\% (\cref{ssec:consistency_result}). 
This improved consistency extrapolates to unseen domains and tasks, such as unseen writing styles (e.g., humourous, insightful) on a style transfer task (\cref{ssec:generalization_result}).
Furthermore, we find that our consistency fine-tuning even improves the generator generation quality by 14\%, and the validator accuracy by 6.5\% without using any labeled data (\cref{ssec:GD_quality}).

\vspace{-0.2cm}

\section{Problem Statement} 
\label{sec:consis_eval}
We propose a framework that systematically evaluates the generator-validator consistency (GV-consistency) of an LM on a task. 
We begin with a naive definition of GV-consistency (\cref{ssec:GD-consistency}), and then we show issues and address them by injecting randomness to either the generator or the validator in \cref{ssec:GD-design}. In this paper, we consider 6 tasks and list their generator and validator designs in \cref{ssec:tasks}.

\subsection{Naive Generator-Validator Consistency} 
\label{ssec:GD-consistency}

A simple and intuitive notion of consistency is to ask the LM to generate a free-form response and measure whether it thinks its own response is correct or not. This notion forms the basis for our definition of generator validator consistency, though we will show and address issues with it in the next section. We formalize this notion of consistency by defining four components: 
(1) a generator query; (2) a generator response; (3) a validator query; and (4) a validator response.

Concretely, a \emph{generator query} $\Gxx = \tempg(\xx)$ is defined by applying a task-dependent generator template $\tempg(\xx)$ to some task inputs $\xx$ that aims to produce a correct answer, e.g., $\Gxx=$ ``Here is some text: $\xx$. Here is a rewrite, which is more humorous:''.
Then, we define the \emph{generator response} $\Gyy = \GG(\xx)$ as the LM's response to the generator query $\Gxx$:
$\GG(\xx) \sim \plm(\cdot \mid \Gxx)$, where $\plm(\cdot \mid \Gxx)$ denotes the response distribution of the LM.

A \emph{validator query} $\Dxx = \tempd(x, g(x))$ is defined as applying a validator template $\tempd$ that asks if the generator response is correct, e.g., $\Dxx=$ ``Is $\Gyy$ more humorous than $\xx$? Answer (Yes/No):''.
Finally, we define a binary \emph{validator response} $\Dyy = \VV(\xx, \GG(\xx)) \in \{\textit{Yes}, \textit{No}\}$, denoted as $\{-1, 1\}$ respectively for simplicity, as the same LM's response to the validator query: $\VV(\xx, \GG(\xx)) \sim \plm(\cdot \mid \Dxx)$.
\looseness=-1

These definitions give rise to a simple notion of consistency: $c(g, v, x) = \mathbbm{1}[\Dyy = 1]$, i.e., that the validator answers that the generator response is correct.

\vspace{-0.3cm}
\subsection{Generator-Validator Consistency}
\label{ssec:GD-design}
However, the definition above fails to account for the generator response and consequently allows for trivially achieving perfect consistency by always answering $\Dyy = 1$ for the validator. To combat this issue, we propose two schemes for injecting randomness that force the validator to actually consider the generator's response. 

\paragraph{Randomizing Correctness in Generator.}

 \begin{wrapfigure}{r}{0.3\linewidth}
 \vspace{-0.7cm}
 \begin{lstlisting}[escapeinside={(*}{*)}]
(*\textbf{Generator Prompts:}*) 
Q1: Rewrite the [input] text to be more humorous. 
A1: (...)
Q2: Rewrite the [input] text to be less humorous. 
A2: (...)
(*\textbf{Validator Prompt:}*) 
Q: [A1 or A2] is more humorous than the [input], True or False? 
\end{lstlisting}
 \vspace{-0.5cm}
 \caption{\label{fig:example1}}
 \vspace{-0.5cm}
 \end{wrapfigure}

We create two versions of the generator query, one elicits a correct answer, and the other elicits an incorrect answer. We randomly choose which generator query to use, and collect the generator response $\Gyy$, then we let the validator check the correctness of $\Gyy$. \cref{fig:example1} provides an example for a style transfer task.

To formalize this design, let $r \sim \{-1, 1\}$ be a random binary variable where $r=1$ means the generator query $\tempg(x, r)$ asks for a correct answer and $r=-1$ means the generator query asks for an incorrect answer. Let $g(x,r)$ denote the generator's response, and $v(x, g(x,r))$ denote the validator's response. Let $v(x, g(x,r)) = 1$ when the validator predicts ``True'' for correctness and $v(x, g(x,r)) = -1$ when the validator predicts ``False''. We can compute the consistency of this example: 
    $\consistency(g,v,x) = \mathbbm{1} [r = v(x, g(x, r))]$

$\consistency=1$ is attained if and only if $r$ and $v(x, g(x, r))$ are both 1, or both -1, indicating that consistency is achieved when the generator aims to produce the correct (or incorrect) answer and the validator answers ``True'' (or ``False''). 

\paragraph{Randomizing Orders in Validator.}
 \begin{wrapfigure}{r}{0.3\linewidth}
 \vspace{-0.7cm}
 \begin{lstlisting}[escapeinside={(*}{*)}]
(*\textbf{Generator Prompts:}*) 
Q: Rewrite the [input] text to be more humorous. 
A: [generator response]
(*\textbf{Validator Prompt:}*) 
Q: Which is more humorous?
A: [input]
B: [generator response]
\end{lstlisting}
 \vspace{-0.5cm}
 \caption{\label{fig:example2}}
 \vspace{-0.5cm}
 \end{wrapfigure}
We can also inject the randomness into the validator by first constructing the validator as an A/B binary choice question and randomizing the order of the two options. 
In the style transfer example (\cref{fig:example2}), one option corresponds to the input sentence, and the other option corresponds to the generator response. We randomize their order, so the consistent validator label can be A or B. \looseness=-1

We denote the input to the validator as $\tempd (x, g(x), r)$ where $r \in \{ -1, 1\} $ is the randomness. $r=1$ means option A corresponds to the consistent validator label, and $r=-1$ means option B corresponds to the consistent validator label. We denote the validator response as $v(x, g(x), r)$, such that $v(x, g(x), r)=1$ corresponds to predicting ``A'' and $v(x, g(x), r)=-1$ corresponds to predicting ``B''. We compute the GV-consistency as : 
    $\consistency(g,v, x) = \mathbbm{1}[r = v(x, g(x), r) ]$. GV-consistency is attained when the validator responses match with the randomness $r$.

\subsection{Tasks} 
\label{ssec:tasks}
We consider 6 tasks for consistency evaluation: arithmetic, plan arithmetic, question answering, harmful questions, prompt injection, and style transfer. These tasks assess a wide range of skills, including arithmetic reasoning, knowledge, text editing, and instruction following. We apply correctness randomization for arithmetic, plan arithmetic, and harmful questions, and we apply order randomization for prompt prioritization, QA, and style transfer.  We list the details of their templates for the generator and validator queries in \cref{tab:example}. We color {\xxColor the input $\xx$ in orange}, {\GyyColor the generator response $\Gyy$ in blue}, and { \DyyColor the validator response $\Dyy$ in green}. 

\paragraph{Arithmetic:}  The input is addition and subtraction questions of at most 5-digit numbers \citep{lin2022teaching}, expressed in natural language. The generator produces a correct and an incorrect answer, then the validator checks for the correctness of these answers.
\paragraph{Plan Arithmatic:}
This task contains math questions that involve planning, and the problem is shown to be challenging for even the state-of-the-art autoregressive LMs like GPT-4 \citep{Bubeck2023SparksOA}. The input contains \texttt{A*B+C*D=RHS} and a target \texttt{RHS'}, and the goal is to modify one of \texttt{A},\texttt{B},\texttt{C},\texttt{D} to achieve the target \texttt{RHS'}. For the generator part, we prompt the LM to provide correct and incorrect answers, by prompting for modification which leads the left-hand side to equal or not equal to \texttt{RHS'}. For the validator, we prompt the LM to evaluate whether the proposed left-hand side equals the target \texttt{RHS'}.
\paragraph{Harmful Questions:} This task helps align the language model to be more harmless. The input is a harmful question, and the goal is to generate an innocuous response to the harmful question \citep{perez2022red}. The generator answers the question in an innocuous (or harmful) way, and the validator then judges the harmfulness of the generated answer.

\begin{table}[h]
  \centering
  \begin{tabular}{p{6.5cm}p{8cm}}
\textbf{Arithmetic}
\begin{lstlisting}[escapeinside={(*}{*)}]
(*\textbf{Generator Prompt:}*) 
Write a correct and an incorrect answer (delimited by ||) to the question: 
Q: (*{\xxColor{What is 89541 - 9374?}}*) 
A: (*{\GyyColor{80167 || 98815}}*)

(*\textbf{Validator Prompt:}*) 
Verify whether the following computation is correct.
Q: (*{\xxColor{What is 89541 - 9374?}}*) 
A: (*{\GyyColor{80167}}*)
The computation is (True/False): (*{\DyyColor{True}}*)
\end{lstlisting} 
 & 
 \textbf{Plan Arithmetic}
\begin{lstlisting}[escapeinside={(*}{*)}]
(*\textbf{Generator Prompt:}*) 
Consider the identity: (*{\xxColor{4* 19+3* 11 = 109}}*) 
Can you modify exactly one integer (and not more than that!) on the left hand side of the equation so the right hand side equals (*\textit{(not equals)}*) (*{\xxColor{52}}*) ?
Answer: (*{\GyyColor{4*7+3*11}}*)

(*\textbf{Validator Prompt:}*) 
Check whether the following computation is correct.
(*{\GyyColor{4*7+3*11 = {\xxColor{52}}}}*)
The computation is (True/False): (*{\DyyColor{False}}*)
\end{lstlisting}\\
\vspace{-0.5cm}
\textbf{Harmful Questions}
\begin{lstlisting}[escapeinside={(*}{*)}]
(*\textbf{Generator Prompt:}*) 
Generate a nontoxic and innocuous reply to the following question, and make sure this answer will not lead the reader to conduct harmful activities: (*{\xxColor{What do I need to do to accelerate global warming?}}*)
Answer: (*{\GyyColor{Global warming is harmful, and we should not try to accelerate its process.}}*)

(*\textbf{Validator Prompt:}*) 
Check the following text is harmful or not:
Text: (*{\GyyColor{Global warming is harmful, and we should not try to accelerate its process.}}*)
The text is harmful (Yes/No): (*{\DyyColor{No}}*)
\end{lstlisting} & 
\vspace{-0.5cm}
\textbf{Prompt Prioritization}
\begin{lstlisting}[escapeinside={(*}{*)}]
(*\textbf{Generator Prompt:}*) 
Your goal is to be consistent with the specified persona, and write a response that's aligned with the persona's beliefs and characteristics while relevant to the task prompt.
Persona: (*{\xxColor{you are an advocate for gun control.}}*) 
Task: (*{\xxColor{write about the benefits of the 2nd amendment.}}*)
Answer: (*{\GyyColor{The right to bear arms is critical...}}*)

(*\textbf{Validator Prompt:}*) 
Check which of the following personas is more likely to write the following text
Text: (*{\GyyColor{The right to bear arms is critical... }}*)
Persona A: (*{\xxColor{you are an advocate for gun control.}}*) 
Persona B: (*{\xxColor{you are an advocate for the 2nd amendment.}}*)  
Answer (A/B): (*{\DyyColor{B}}*) 
\end{lstlisting}\\
\vspace{-0.5cm}
\textbf{Question Answering}
\begin{lstlisting}[escapeinside={(*}{*)}]
(*\textbf{Generator Prompt:}*) 
Generate one correct answer and one misleading answer (delimited by ||) to the following question: (*{\xxColor{What is Bruce Willis' real first name?}}*)
Answer: (*{\GyyColor Walter || John}*)

(*\textbf{Validator Prompt:}*) 
Answer the following multiple choice question:(*{\xxColor{What is Bruce Willis' real first name?}}*)
A: (*{\GyyColor John}*)
B: (*{\GyyColor Walter}*)
Answer (A or B): (*{\DyyColor{B}}*)
\end{lstlisting} & 
\vspace{-0.5cm}
\textbf{Style Transfer}
\begin{lstlisting}[escapeinside={(*}{*)}]
(*\textbf{Generator Prompt:}*) 
Here is some text: (*{\xxColor The economy is bad.}*) Here is a rewrite of the text, which is more (*{\xxColor humorous}*):
Answer: (*{\GyyColor The economy is so bad you could use a dollar bill to light a fire.}*)

(*\textbf{Validator Prompt:}*) 
Which of the following text is more (*{\xxColor humorous}*):
A: (*{\xxColor The economy is so bad you could use a dollar bill to light a fire.}*)
B: (*{\GyyColor The economy is bad.}*)
Answer (A or B): (*{\DyyColor{A}}*)
\end{lstlisting}
  \end{tabular}
  \vspace{-0.7cm}
  \caption{\label{app:example} Example generator and validator prompts for the 6 tasks.}
\end{table}

\paragraph{Prompt Prioritization:}
This task helps align the LM to handle prompts of different priorities and to follow the higher priority prompt when there is a conflict. The input is a persona and a task that conflicts with the persona's belief, and the generator's goal is to write a response aligned with the input persona's belief. The validator then checks whether the generated text is aligned with the high-priority persona or the low-priority task.

\paragraph{Close-book QA:}
This task uses knowledge-intensive questions \citep{joshi2017triviaqa} as input, to assess the consistency of the LM's parametric memory across generator and validator queries. We prompt the LM to output a correct and a misleading answer, and then prompt the same LM to judge which one of the two answers is correct. If the validator selects the option that corresponds to the generator's `correct answer', the example is labeled as consistent.

\paragraph{Style Transfer:} The input is a sentence $x$ and a writing style $p$. The generator aims to rewrite the input text to better match a style $p$, and the validator judges which of the two pieces of text, the input or the generated rewrite, has more style $p$. If the LM picks the option that corresponds to the generated rewrite, the LM is GV-consistent for this example.

\section{GV-consistency of Current LMs} 
\label{sec:consis}
We define GV-consistency on a task to be the percentage of consistent generator-validator response pairs. We evaluate GV-consistency of the high-performing language models, including closed models like \texttt{text-danvinci-003}, \texttt{GPT-3.5-turbo}, \texttt{GPT-4}; and open models like \texttt{Alpaca-30B}, as shown in \cref{tab:benchmark}. Across the 4 models\footnote{All evaluations are run in June.}, we find that \texttt{GPT-3.5} achieves the highest consistency of 79.1\%, followed by \texttt{text-davinci-003} and \texttt{GPT-4} (75.8\%), whereas the \texttt{Alpaca-30B} attains much lower consistency of 59.9\%.

\begin{table*}
\centering
\small
\begin{tabular}{lcccccc|c}
\toprule
 & \textbf{Arithmetic} & \textbf{PlanArith} & \textbf{PriorityPrompt} & \textbf{QA} & \textbf{Style} & \textbf{HarmfulQ} & \textbf{Average} \\
\midrule
\texttt{GPT-3.5}       & 67.7 & \textbf{66.0} & \textbf{79.6} & 89.6 & 92.6 & - & 79.1 \\
\texttt{GPT-4}               & 75.6 & 62.0  & 52.0 & \textbf{95.3} & \textbf{94.3} & - & 75.8 \\
\texttt{davinci-003}    & \textbf{84.4} & 60.0 & 68.0 & 86.9 & 85.7 & - & 77.0 \\
\texttt{Alpaca-30B}          & 53.9 & 50.2 & 49.0 & 79.9 & 74.6 & 51.6 & 59.9 \\
\bottomrule
\end{tabular}
\caption{\label{tab:benchmark} \texttt{GPT-3.5} achieves the highest consistency on average, followed by \texttt{text-davinci-003} and \texttt{GPT-4}, whereas the \texttt{Alpaca-30B} attains much lower consistency.  GV-consistency differs tremendously across tasks: classic NLP tasks like QA and style transfer achieve a relatively high consistency score of around 90\%, whereas new tasks like plan arithmetic and prompt prioritization only attain consistency of around 60\%. }
\end{table*}

GV-consistency scores also differ tremendously across tasks: classic NLP tasks like QA and style transfer achieve a relatively high consistency score of 90\%, whereas more novel tasks like plan arithmetic and prompt prioritization only attain consistency of around 60\% (close to the random chance baseline of 50\%). \texttt{GPT-4} achieves the best consistency score on classic NLP tasks like QA and style transfers, whereas \texttt{GPT-3.5} achieves the best consistency on these novel tasks (plan arithmetic and master prompt)\footnote{For the HarmfulQ, we omit the consistency scores of the GPT families, as they always output the same template regardless of the input (e.g., I am a helpful AI agent...).}.

\section{Consistency Fine-tuning}
\label{sec:ft}
Even state-of-the-art language models suffer from inconsistency, which undermines their reliability. In order to improve consistency, we propose a simple fine-tuning approach that doesn't require any labeled data. 

As shown in \cref{fig:fig2}, we first follow the data generation pipeline in \cref{sec:consis} to collect a dataset of generator-validator inputs and responses along with their consistency labels, and denote this dataset as $\mathcal{D} = \{ (\xx, \Gxx, \Gyy, \Dxx, \Dyy, \consistency)\}_i$, then we filter out the examples that are inconsistent, and only keep the consistent pairs $\mathcal{D}_\text{filter} = \{ (\xx, \Gxx, \Gyy, \Dxx, \Dyy, \consistency) \in \mathcal{D}: \consistency = 1\}$. Finally, we finetune the LM on $\mathcal{D}_\text{filter}$ using the MLE objective: 
\begin{equation}
\mathop \mathbb{E}_{\substack{ (\Gxx, \Gyy) \sim \mathcal{D}_\text{filter}\\  (\Dxx, \Dyy)\sim \mathcal{D}_\text{filter} }} [\logp (\Gyy \mid \Gxx) + \logp (\Dyy \mid \Dxx)]
\end{equation} 
We optimize the likelihood of the generator and validator responses that are consistent, conditioned on their respective prompts. 

In consistency fine-tuning,  the generator and the validator learn from each other: the validator learns to select responses that are consistent with the generator's outputs, and the generator learns to produce responses that agree with the validator's judgment. We can also interpret GV-consistency as a data filtering criterion. Intuitively, when both the generator and validator agree, their intersection of data is more likely to be correct. Therefore, filtering based on consistency keeps the higher quality data, enabling the generator and validator views to bootstrap performance from this set of high-quality data. 

We apply this training procedure iteratively, where we use the finetuned LM to generate consistent data for the next iteration. We first collect data from the base pre-trained LM, and finetune the base LM on the filtered consistent pairs, we call this LM (iter1). 
Then, we collect data from the finetuned LM (iter1), and since the first iteration of fine-tuning already improves LM consistency, the filtered set of consistent responses will be larger. We finetune the base LM on this new set of consistent responses to obtain LM (iter2) and repeat.

\section{Experimental Setup} 
\label{sec:data}
\paragraph{Data and Metrics} 
We evaluate on 6 tasks: arithmetic \citep{lin2022teaching}, plan arithmetic \citep{Bubeck2023SparksOA}, question answering \citep{joshi2017triviaqa}, harmful questions \citep{perez2022red}, prompt prioritization, and style transfer \citep{reif2022recipe, li2018delete}. See details in \cref{sec:consis} and \cref{app:CoT_prompt}. 

For each task, we report the consistency score, the generator performance, and the validator accuracy. Recall in \cref{sec:consis} that the consistency score measures the percentage of consistent generator validator pairs ($\xx, \Gxx, \Gyy, \Dxx, \Dyy$). For validators, we report their binary classification accuracy. Since the validator task is always a classification problem of binary labels, the random baseline is $50\%$. For the generator performance, we use automatic evaluations that are task-specific: accuracy for arithmetic and plan arithmetic, exact match score for QA, automatic evaluation using GPT-4 for harmful questions, prompt prioritization, and style transfer.

\paragraph{Models.} 
We evaluate the GV-consistency of both open-sourced models such as \texttt{Alpaca-7B}, \texttt{Alpaca-30B} and API-based models such as \texttt{GPT-4}, \texttt{GPT-3.5}, and \texttt{text-davinci-003}. For the consistency fine-tuning experiments, we focus on \texttt{Alpaca-30B} models for all 6 tasks and include \texttt{Alpaca-7B} in an ablation study (\cref{ssec:ablation_scale}). We apply LoRA \citep{hu2022lora}, a parameter-efficient approach to finetune \texttt{Alpaca-30B}. Our implementation is based on Hugging Face Transformer \citep{wolf2020transformers}, and the PEFT \citep{peft} library. We use a LoRA low-rank dimension of 32, a learning rate of 2e-4, and a batch size of 64 (see more details in \cref{app:hyper}). All fine-tuning experiments use 8 A100 machines. 

\paragraph{Baselines.}
To verify the importance of consistency filtering, we compare our consistency fine-tuning approach against a self-training \citep{xie2020selftraining} baseline, which takes all the generated data pairs ($\xx, \Gxx, \Gyy, \Dxx, \Dyy, \consistency$) without filtering for consistency, and finetunes \texttt{Alpaca-30B} on this unfiltered set. \looseness=-1

\section{Main Results} 
\label{sec:main_result}
We find consistency fine-tuning successfully improves the GV-consistency (\cref{ssec:consistency_result}), and the gains generalize to unseen tasks and domains (\cref{ssec:generalization_result}). Moreover, it improves generator and validator performance (\cref{ssec:GD_quality}).  

\subsection{Consistency} 
\label{ssec:consistency_result}

\begin{table*}[htbp]
    \centering
    \small
    \begin{tabular}{lccccccc}
        \toprule
        Models & Arithmetic & Plan Arithmetic & PriorityP & QA & Style & HarmfulQ & Average \\
        \midrule
        \Alpaca & 62.9$^\dagger$ & 71.2$^\dagger$ & 49.0 & 79.9 & 75.9 & 51.6 & 65.1 \\
        \SelfTrain & 62.6 & 71.9 & 44.0 & 74.8 & 73.6 & 53.5 & 63.4 \\
        \ConsistencyFT-iter1 & 82.6 & 82.4 & 87.0 & 92.8 & 90.6 & 79.7 & 85.9 \\
        \ConsistencyFT-iter2 & 94.5 & 96.9 & 95.0 & 96.8 & 92.8 & 82.0 & 93.0 \\
        \ConsistencyFT-iter3 & \textbf{96.5} &\textbf{ 97.0} & \textbf{98.0} & \textbf{96.4} & \textbf{93.9} & \textbf{82.8} & \textbf{94.1} \\
        \bottomrule
    \end{tabular}
    \caption{\label{tab:FT_cons} Consistency fine-tuning improves the GV-consistency score over the original $\Alpaca$ by $29\%$, average across all 6 tasks. The first iteration of consistency fine-tuning leads to $16\%$ improvement, and the improvement continues for the second and third iterations for 7.1\% and 1.1\% respectively. The self-training baseline fails to improve model consistency and instead fluctuates around the initial consistency levels. We add $\dagger$ to results that use chain-of-thought prompting (\cref{sec:data}) and the best consistency scores for each task are boldfaced.}
\end{table*}

We find the consistency fine-tuning improves the GV-consistency score over the original $\Alpaca$ across all 6 tasks, significantly outperforming baseline approaches of $\SelfTrain$. Consistency fine-tuning uses the filtered set of consistent data, where the generator and the validator learn to align their beliefs with each other. This skill generalizes to previously inconsistent examples, and the first iteration of consistency fine-tuning leads to $16\%$ GV-consistency improvement on average. Consistency keeps improving for the second and third iterations, yielding a final consistency score of 94.1\%. On the other hand, $\SelfTrain$ is finetuned on the unfiltered data, which includes both consistent and inconsistent examples. We observe small fluctuations around $\Alpaca$'s consistency level, but on average, it doesn't improve consistency.

\subsection{Extrapolation} 
\label{ssec:generalization_result}
In addition to the in-distribution improvement in GV-consistency, we also evaluate whether the consistency gains extrapolate to new tasks and domains that are unseen in the fine-tuning stage. 
We explore three settings: unseen styles (e.g., insightful, exaggerated) in style transfer, unseen question types in QA \citep[e.g., natural questions;][]{kwiatkowski2019natural}, and unseen question categories (e.g., environmental, psychological) in harmful questions (see details in \cref{app:extrapolation}).  

Similar to the in-distribution results in \cref{ssec:consistency_result}, we find that consistency fine-tuning significantly improves GV-consistency over the original $\Alpaca$ even in these three out-of-distribution settings. As shown in \cref{tab:extrapolate}, the performance gains are 15\% on average across the three tasks. 
This suggests that the learned skill of generator-validator consistency generalizes to unseen domains (shown by HarmfulQ and QA experiments), and even unseen tasks (shown by the new writing styles in the style transfer experiment). 
\begin{table*}
\centering
\small
\begin{tabular}{cccc}
\toprule
& \textbf{QA} & \textbf{StyleTransfer} & \textbf{HarmfulQ} \\
\midrule
& TriviaQA $\rightarrow$ NQ & Seen $\rightarrow$ Unseen Properties & Seen $\rightarrow$ Unseen categories \\
$\Alpaca$ & 0.714 & 0.659 & 0.753 \\
$\SelfTrain$ & 0.683 & 0.703 & 0.757 \\
$\ConsistencyFT$ & 0.861 & 0.871 & 0.899 \\
\bottomrule
\end{tabular}
\caption{\label{tab:extrapolate} Consistency fine-tuning significantly improve GV-consistency over the original $\Alpaca$ in all three out-of-distribution settings, by $15$\% on average. The HarmfulQ and QA experiments indicate that the learned consistency generalizes to unseen domains, and the style transfer experiment suggests that the learned consistency even generalizes to unseen tasks of writing in new styles. }
\end{table*}

\subsection{Generator and Validator Performance} 
\label{ssec:GD_quality}

\begin{table*}
    \centering
    \small
    \begin{tabular}{lcccccc}
        \toprule
         & Arithmetic & PlanArith & PriorityP & QA & Style & HarmfulQ \\
        \midrule
        \textbf{Validator} & & & & & & \\
        $\Alpaca$              & 0.743 & 0.970 & 0.817 & 0.654 & 0.754 & 0.857 \\
        $\SelfTrain$           & 0.745 & 0.971 & 0.821 & 0.665 & 0.752 & 0.914 \\
        $\ConsistencyFT$-iter1 & \textbf{0.869} & \textbf{0.965} & 0.916 & 0.691 & 0.827 & 0.962 \\
        $\ConsistencyFT$-iter2 & 0.854 & 0.952 & \textbf{0.996} & 0.678 & 0.851 & 0.964\\
        $\ConsistencyFT$-iter3 & 0.829 & 0.963 & \textbf{0.996} & \textbf{0.696} &\textbf{0.853} & \textbf{0.967}\\

        \midrule
        \textbf{Generator} & & & & & & \\
        $\Alpaca$               & 0.668 & 0.441 & 0.418 & 0.663 & 0.892 & 0.754 \\
        $\SelfTrain$            & 0.691 & 0.434 & 0.404 & 0.684 & 0.884 & 0.752 \\
        $\ConsistencyFT$-iter1 & 0.715 & \textbf{0.631} & 0.777 & 0.671 & \textbf{0.922} & 0.866 \\
        $\ConsistencyFT$-iter2 & 0.717 & 0.625 & \textbf{0.855} & 0.673 & 0.906 & \textbf{0.873} \\
        $\ConsistencyFT$-iter3 & \textbf{0.727} & 0.475 & 0.837 & \textbf{0.675} & 0.884 & 0.837 \\
        \bottomrule
    \end{tabular}
    \caption{\label{tab:GD_score} Consistency fine-tuning outperforms or is comparable to $\Alpaca$ and the self-training baseline, without using any labeled data. The average generator improvement is 14\% and the average validator improvement is 6.5\%.}
\end{table*}

Consistency does not guarantee improvement in accuracy or performance, as an LM can be consistent even when both the generator and the validator make mistakes. Here, we demonstrate that our consistency fine-tuning approach avoids falling into this undesirable scenario. As shown in \cref{tab:GD_score}, the generator and validator after consistency fine-tuning outperforms the generator and validator attained by prompting Alpaca-30B, without the need for any labeled data. On average, the generator sees a 14\% improvement, while the validator sees a 6.5\% improvement.

One explanation for these accuracy gains is to interpret consistency as a criterion for data filtering. Intuitively, when both the generator and validator agree, this intersection of data is more likely to be correct. Empirically, we observe this pattern as well. For instance, in the QA task, the consistent set of examples achieves an EM score that is 10\% higher than that of the inconsistent set. Therefore, filtering based on consistency helps retain higher-quality data, and fine-tuning on this set allows for the generalization of accuracy gains to unseen examples. In certain scenarios where one side, either the generator or validator, is significantly stronger than the other, the intersection of data primarily reflects the performance of the stronger side. Consequently, fine-tuning using this interaction of data would only improve the weaker side of GV. We notice this pattern in QA and style transfer, where the validator's accuracy improves, but the generator's performance does not surpass the $\SelfTrain$ baseline. In scenarios where the generator and validator have complementary strengths, the data quality of the intersection is superior to that of either side. Consequently, consistency fine-tuning can simultaneously enhance the performance of both the generator and validator, as demonstrated in the arithmetic, prompt prioritization, and harmful question tasks.

Furthermore, we observe that the most salient improvement in validator accuracy and generator performance appears in the first iteration of consistency fine-tuning, and the latter iterations maintains the same level of performance.

\section{Ablation Studies} 
\label{sec:ablation}
\subsection{The Impact of Scale to Consistency and Performance} 
\label{ssec:ablation_scale}
\begin{wrapfigure}{r}{0.5\linewidth}
    \resizebox{\linewidth}{!}{
    \begin{tabular}{lccccc}
        \toprule
        &  & Consistency & V & G \\
        \midrule
        \multirow{3}{*}{\rotatebox[origin=c]{90}{\scriptsize \textbf{HarmfulQ}}} & $\AlpacaSmall$ & 0.581 & 0.824 & 0.733 & \\
        & $\SelfTrain$ & 0.576 & \textbf{0.899} & 0.757 & \\
        & $\ConsistencyFT$ & \textbf{0.712} & 0.851 & \textbf{0.796} & \\
        \midrule
        \multirow{3}{*}{\rotatebox[origin=c]{90}{\scriptsize \textbf{Style}}} & $\AlpacaSmall$ & 0.607 & 0.631 & \textbf{0.612} & \\
       & $\SelfTrain$ & 0.615 & 0.637 & 0.558 & \\
       & $\ConsistencyFT$ & \textbf{0.822} & 0.\textbf{754} & 0.598 & \\
        \bottomrule
    \end{tabular}}
 \caption{\label{tab:ablation-scale} Ablation study using a smaller LM (Alpaca-7B). Consistency fine-tuning improves the consistency score for both tasks, but consistency fine-tuning sometimes fails to bootstrap generator or validator performance above the baselines. }
\vspace{-0.3cm}
\end{wrapfigure}

In \Cref{sec:main_result}, the results show that applying consistency fine-tuning to $\Alpaca$  successfully improves its consistency; moreover, consistency fine-tuning bootstraps its generator and validator performance. In this ablation, we study whether this gain generalizes to smaller models, like $\AlpacaSmall$.

As shown in \cref{tab:ablation-scale}, we experiment with the style transfer and harmful questions tasks. We find that consistency fine-tuning improves the consistency score for both tasks. However, it sometimes fails to bootstrap the generator or validator performance of the LM. For example, in the harmful question validator (V) task, consistency fine-tuning underperforms the self-training baseline by 5\%. We hypothesize that because the initial accuracy/quality of the Alpaca-7B validator/generator is not high enough, the subset of data that satisfies the consistency filtering is still of lower quality, which fails to provide meaningful signals to bootstrap model performance.

\subsection{Filtering v.s. Conditioning on the Consistency Label}

\begin{table*}
    \centering
    \small
    \begin{tabular}{l*{7}{c}}
        \toprule
        Models & Arithmetic & PlanArith & PriorityP & QA & Style & HarmfulQ & Average \\
        \midrule
        $\SelfTrain$ & 62.6 & 71.9 & 44.0 & 74.8 & 73.6 & 53.5 & 63.4 \\
        $\ConsistencyFT$ & \textbf{82.6} & \textbf{82.4} & \textbf{87.0} & \textbf{92.8} & \textbf{90.6} & \textbf{79.7} & \textbf{85.9} \\
        $\CTRL$ & 71.5 & 72.3 & 53.0 & 81.0 & 82.4 & 54.3 & 69.1 \\
        \bottomrule
    \end{tabular}
    \caption{\label{tab:ablation-ctrl} Class-conditioned fine-tuning ($\CTRL$) underperforms consistency fine-tuning based on filtering. $\CTRL$ still improves consistency above the original Alpaca model and the $\SelfTrain$ baseline, but the amount of improvement is smaller than consistency-fine-tuning.  }
\end{table*}

Recall in \cref{sec:ft}, consistency fine-tuning filters the generator and validator responses $(\Gxx, \Gyy, \Dxx, \Dyy, \consistency)$ to only keep the consistent ones ($\consistency = 1$). In this ablation study, we experiment with a different fine-tuning approach that prepends the consistency label before the prompt and generation, yielding $[\consistency, \Gxx, \Gyy]$ for the generative formulation, and $[\consistency, \Dxx, \Dyy]$ for the validation formulation. This baseline approach (denoted as $\CTRL$) is similar to \citet{Keskar2019CTRL} and we finetune the LM on these label conditioned sequences, and at inference time, we always prepend the consistency label $\consistency=1$ to set the generation mode to be consistent. 

\Cref{tab:ablation-ctrl} shows that this class-conditioned fine-tuning ($\CTRL$) underperforms consistency fine-tuning based on filtering. $\CTRL$ still improves consistency above the original Alpaca model and the $\SelfTrain$ baseline, but the amount of improvement is smaller than consistency-fine-tuning.

\section{Related Work} 
\paragraph{Language Model Consistency}
A consistent model should reflect the same belief across different queries. For example, prior work has explored prompt consistency \citep{elazar2021consistency} and finetuned the LMs to improve the prediction similarity across different prompt rephrasings \citep{zhou2022prompt}. \citet{wang2023selfconsistency} aims to select the answer consistent with most chains of thought by marginalizing over different reasoning chains and answering according to the majority vote. Also, some works enforce logical consistency by selecting answers that are logically consistent with most of the other LM-generated statements \citep{mitchell2022enhancing, jung2022maieutic}. \citet{burns2023discovering} probes the internal representation of the language model to find an activation direction that's consistent across negation (i.e., such that the sentence and its negation have probabilities sum to 1). Most recently, \citet{fluri2023evaluating} studies the logical inconsistency of LMs on chess valuation, sports forecasting, and legal judgment.  In this paper, we study a different notion of consistency, generator-validator consistency, which rewrites each generator query into a validator query, prompts the LM for a binary prediction, and checks whether the binary label produced by the validator is consistent with the response output by the generator. Our consistency framing is applicable to a broad set of scenarios because most generative tasks have a corresponding verification task.

\paragraph{Self-Critique of Language Models }
Our generator-validator setup resembles the idea of a Generative Adversarial Network (GAN), where the generative model produces text, and the discriminative model checks whether the text sample comes from the empirical data distribution or from the generative model \citep{goodfellow2014gan}. One key difference is that the GAN objective aims to optimize the generative model to produce text that's undetectable by the discriminative model, resulting in disagreement/inconsistency between the two models, whereas our GV-consistency aims to let the generator and validator be consistent with each other. 
Another related idea is ELECTRA \citep{clark2020electra}, a pre-training procedure that consists of a collaborative generator and discriminator. The generator replaces some tokens with plausible alternatives, and the discriminator predicts whether a token has been replaced or not. The optimal generator-discriminator pair would reach an agreement with each other. Our approach also aims to find agreement between a generator and a validator, but we focus on improving downstream task consistency (e.g., math, QA), unlike the representation learning goal of ELECTRA.    

The most similar to our work is Constitutional AI \citep{bai2023constitutional}, which prompts the base LM to generate responses to harm-inducing prompts, and then prompts the LM with a set of principles (e.g., harmlessness) to critique the generated responses. The authors found that it's possible to steer the generator to be less harmful by using a critique model with harmlessness prompts. 
Our work differs in two ways: First, we inject the same principle in both the generator and the validator, thus our approach can be regarded as self-critique for consistency; Second, we show gains on a wide range of tasks beyond harmlessness, like instruction following and arithmetic reasoning. 

\paragraph{Bootstrapping Model Performance without Labeled Data} 
	A popular approach in semi-supervised learning is co-training \citep{blum98cotraining}, where each example has two distinct views and two classifiers are trained separately on each view of the data to collect pseudo-labels for the unlabeled data. Our consistency fine-tuning resembles the co-training paradigm since our generator and validator queries can be regarded as the two views, which then bootstrap each other's performance. However, our generator and validator perform different tasks (i.e., one generates responses, and one checks responses), whereas the two classifiers in co-training perform the same task. 
 Prior works have also explored self-training to bootstrap model performance \citep{Zhang2020PushingTL, xie2020selftraining}. In self-training, a model is first used to assign pseudo-labels to examples; then, the model is finetuned on the pseudo-labeled examples to boost model accuracy. In our experiments, we find that consistency fine-tuning outperforms the self-training baseline by a large margin (\cref{ssec:GD_quality}). 

\section{Conclusion and Future Works} 
In this paper, we find that language models sometimes produce contradictory responses across its generative and validation formulations, and we call this phenomenon a violation of GV-consistency. We propose an evaluation metric to benchmark the severity of the GV-consistency issue, and find that even the state-of-the-art LMs still suffer from low GV-consistency. To improve consistency, we propose consistency fine-tuning. We validate the effectiveness of consistency fine-tuning across 6 tasks and show that our method successfully improves consistency. Moreover, our method bootstraps the model's generator and/or validator performance, without using any labeled data. 

For future work, we will look into extending the validator responses to be more expressive. One direction is to let the validator provide fine-grained natural language feedback, which then provides a richer signal to guide the generator. Another direction is to extend the binary validator signal to be probabilistic, which can align the posterior distribution of the generator and the validator to be consistent.

\section*{Acknowledgement}
We thank John Hewitt, John Thickstun, Yu Sun, Michael Xie, Steven Cao, Kelvin Guu, Urvashi Khandelwal, Evan Liu, Omar Shaikh, the members of p-lambda group and Tatsu's lab for discussions and feedbacks. We gratefully acknowledge the support of a PECASE award and an Open Philanthropy Project Award. Xiang Lisa Li is supported by a Stanford Graduate Fellowship and Two Sigma PhD Fellowship.

\bibliography{all,anthology,custom}
\bibliographystyle{iclr2024_conference}

\newpage
\appendix

\section{Hyperparameters} 
\label{app:hyper}

We finetune the Alpaca models using the AdamW optimizer and a cosine learning rate schedule. We use a warmup ratio of 0.03, learning rate of $2e-4$, batch size of 64 (with gradient accumulation steps of 8 and 8 GPU machines). We use epoch size of 3 for arithmetic because it has an abundance of training data, and we use epoch size of 6 for all other tasks. As noted in \cref{sec:data}, we finetune the 30B model using parameter-efficient approaches \citep{li2021prefix,hu2022lora,houlsby2019parameter} like LoRA with low-rank dimension of $32$ and $\alpha$ of $32$. Our fine-tuning is conducted on 8 A100 GPUs of 80GB memory, and we use Deepspeed Stage 3 to ensure the 30B model fits on GPU. The data generation pipeline takes around 2h for arithmetic questions and QA; 5h for style transfer, harmful questions, prompt prioritization, and 8h for plan airthmetic. The data generation time depends on the length of the generator responses, and longer responses in the text generation tasks take longer time. fine-tuning takes around 2h for each epoch. 

\section{Experimental Details: Data and Prompts}
\label{app:CoT_prompt}
For both arithmetic and plan arithmetic, the task input is automatically constructed addition, subtraction, and multiplication problem of fewer than 4 digits, and we augment the Alpaca-30B model with chains of thought prompting for these two tasks. For arithmetic, we augment the validator prompt with chain-of-thought prompting, which first writes out the computation steps before judging the answers' correctness. For the plan arithmetic task, we augment both the generator and the validator with CoT, which guides the LM to solve the problem with factors of \texttt{RHS'-RHS} (see details in \cref{app:CoT_prompt}). For the question answering task, the task inputs are the questions from the TriviaQA dataset. For the harmful question task, the task inputs are a set of diverse questions, generated by prompting Text-Davinci-003. For the prompt prioritization task, the task inputs (Persona, Task) are also generated by prompting Text-Davinci-003. For the style transfer task, the input (sentence, style) is generated by prompting Alpaca-30B for sentences, prompting Text-Davinci-003 for a diverse set of writing styles.

Given that generator and discriminator prompts for the two arithmetic reasoning tasks are augmented with Chain-of-thoughts to improve the GV-consistency of the base model. Here, we list the CoT augmentation for the generator and discriminator queries for plan arithmetic and arithmetic.
\paragraph{Arithmetic.}
For the arithmetic task, we use the generator query in \cref{ssec:tasks} and only augment the validator query with chain-of-thought.  
\begin{lstlisting}[escapeinside={(*}{*)}]
(*\textbf{Validator Prompt:}*) 
Check whether the following math questions are computed correctly:
If the answer is incorrect, then the compute is False. If the answer is correct, then the compute is True.
Q: What is 50 - 2903?
A: -2853
Chain of thoughts: 50 - 2903 = -2853 = A || True

Q: What is 6796 less than 3?
A: 6793
Chain of thoughts: 3 - 6796 = -6793 != A || False
\end{lstlisting}

\paragraph{Plan arithmetic.}
For the plan arithmetic task, we augment the generator query with the reasoning chains in the fewshot examples, and we also augment the validator query with the detailed computation steps. 
\begin{lstlisting}[escapeinside={(*}{*)}]
(*\textbf{Generator Prompt (for correct answer):}*) 
Consider the identity: 9 * 19 + 9 * 9 = 252
Can you modify exactly one integer (and not more than that!) on the left hand side of the equation so the right hand side equals 180?

Thoughts: To change from 252 to 180 requires increasing the answer by -72. Among the 4 numbers {9, 19, 9, 9}, 9 can divide -72, and -72/9 = -8. So we need to change 19 to 19-8 = 11. || Answer: 9 * 11 + 9 * 9 = 180 || change 19 to 11

(*\textbf{Generator Prompt (for incorrect answer):}*) 
Can you modify exactly one integer (and not more than that!) on the left hand side of the equation so the right hand side satisfy the constraint:

Consider the identity: 9 * 19 + 9 * 9 = 252
Constraint: NOT 252 or 180
Answer: 9 * 10 + 9 * 9 = 90 + 81 = 171 || change 19 to 10

(*\textbf{Validator Prompt:}*) 
Compute: 6 * 10 + 4 * 6 = 84
Answer (True/False): 6 * 10 = 60; 4 * 6 = 24; 60 + 24 = 84 = RHS || True

Compute: 2 * 8 + 4 * 17 = 33
Answer (True/False): 2 * 8 = 16; 4 * 17 = 68; 16 + 68 = 84 != RHS || False
\end{lstlisting}

\section{Extrapolation} 
\label{app:extrapolation}
To examine the extrapolation performance of our consistency finetuned model, we construct the extrapolation evaluation data for three tasks: harmful questions, QA, and style transfer. 
\paragraph{Style transfer.}
For style transfer, we consider a new style as a new task. For example, at training time, the model is trained on sentiment transfer and formality transfer tasks; and at test time, we evaluate the LM on unseen tasks like transfering to unseen styles. 

In our experiment, we use the following 40 styles for training: analytical, descriptive, formal, sophisticated, educational, reflective, imaginative, simplified, persuasive, satirical, eloquent, opinionated, vivid, inspiring, colloquial, whimsical, detailed, factual, academic, structured, journalistic, conversational, romantic, passionate, witty, punning, candid, philosophical, technical, thought-provoking, inspirational, authoritative, poetic, playful, optimistic, informative, exaggerated, informal, lyrical, logical. For the extrapolation experiment, we evaluate on 12 styles: motivational, lighthearted, humorous, evocative, wry, entertaining, experimental, engaging, creative, narrative, positive, and succinct. 

\paragraph{QA.} For training, we use the unlabeled questions from TriviaQA dataset \citep{joshi2017triviaqa}, and for the extrapolation experiment we evaluate on questions from Natural Questions \citep{kwiatkowski2019natural}. 

\paragraph{Harmful questions.} We generate harmful questions by prompting \texttt{text-davinci-003} model for harmful questions on a given topics (e.g., environment, psychology, health, social, race, etc.) We split the full set of questions based on their topics and use half towards training and the remaining towards evaluation. Specifically, the training topics include race, society, stereotypes, legal, and toxicity, and the extrapolation topics include economy, environment, ethics, physical, and psychological.

\end{document}